\documentclass[11pt,letterpaper]{article}
\usepackage{emnlp2016}
\usepackage{times}
\usepackage{latexsym}
\usepackage[T1]{fontenc}

\emnlpfinalcopy

\def\emnlppaperid{349}

\title{Representing Verbs with Rich Contexts: an Evaluation on Verb Similarity}

 \author{Emmanuele Chersoni \\ Aix-Marseille University \\ emmanuelechersoni@gmail.com
        \And Enrico Santus \\ The Hong Kong Polytechnic University \\ esantus@gmail.com
				\AND Alessandro Lenci \\ University of Pisa \\ alessandro.lenci@unipi.it 
				\And Philippe Blache \\ Aix-Marseille University \\ philippe.blache@univ-amu.fr
				\AND Chu-Ren Huang \\ The Hong Kong Polytechnic University \\ churen.huang@polyu.edu.hk}
\date{}
\makeindex
\begin{document}

\maketitle

\begin{abstract}
Several studies on sentence processing suggest that the mental lexicon keeps track of the mutual expectations between words. Current DSMs, however, represent context words as separate features, thereby loosing important information for word expectations, such as word interrelations. In this paper, we present a DSM that addresses this issue by defining verb contexts as joint syntactic dependencies. We test our representation in a verb similarity task on two datasets, showing that joint contexts achieve performances comparable to single dependencies or even better. Moreover, they are able to overcome the data sparsity problem of joint feature spaces, in spite of the limited size of our training corpus.
\end{abstract}

\section{Introduction}

Distributional Semantic Models (DSMs) rely on the Distributional Hypothesis \cite{harris1954distributional,sahlgren2008distributional}, stating that words occurring in similar contexts have similar meanings. On such theoretical grounds, word co-occurrences extracted from corpora are used to build semantic representations in the form of vectors, which have become very popular in the NLP community. Proximity between word vectors is taken as an index of meaning similarity, and vector cosine is generally adopted to measure such proximity, even though other measures have been proposed \cite{weeds2004characterising,santus2016testing}.

Most of DSMs adopt a bag-of-words approach, that is they turn a text span (i.e., a word window or a parsed sentence) into a set of words and they register separately the co-occurrence of each word with a given target. The problem with this approach is that valuable information concerning word interrelations in a context gets lost, because words co-occurring with a target are treated as independent features. This is why works like Ruiz-Casado et al. \shortcite{ruiz2005using}, Agirre et al. \shortcite{agirre2009study} and Melamud et al. \shortcite{melamud2014probabilistic} proposed to introduce richer contexts in distributional spaces, by using entire word windows as features. These richer contexts proved to be helpful to semantically represent verbs, which are characterized by highly context-sensitive meanings, and complex argument structures. In fact, two verbs may share independent words as features despite being very dissimilar from the semantic point of view. For instance \emph{kill} and \emph{heal} share the same object nouns in \emph{The doctor healed the patient} and the \emph{The poison killed the patient}, but are highly different if we consider their joint dependencies as a single context. Nonetheless, richer contexts like these suffer from data sparsity, therefore requiring either larger corpora or complex smoothing processes.

In this paper, we propose a syntactically savvy notion of \textbf{joint contexts}. To test our representation, we implement several DSMs and we evaluate them in a verb similarity task on two datasets. The results show that, even using a relatively small corpus, our syntactic joint contexts are robust with respect to data sparseness and perform similarly or better than single dependencies in a wider range of parameter settings. 

The paper is organized as follows. In Section 2, we provide psycholinguistic and computational background for this research, describing recent models based on word windows. In Section 3, we describe our reinterpretation of joint contexts with syntactic dependencies. Evaluation settings and results are presented in Section 4.

\section{Related Work}

A number of studies in sentence processing suggests that verbs activate expectations on their typical argument nouns and vice versa \cite{mcrae1998modeling,mcrae2005basis} and nouns do the same with other nouns occurring as co-arguments in the same events \cite{hare2009activating,bicknell2010effects}. Experimental subjects seem to exploit a rich event knowledge to activate or inhibit dynamically the representations of the potential arguments. This phenomenon, generally referred to as \textit{thematic fit} \cite{mcrae1998modeling}, supports the idea of a mental lexicon arranged as a web of mutual expectations.

Some past works in computational linguistics \cite{baroni2010distributional,lenci2011composing,sayeed2014combining,greenberg2015improving} modeled thematic fit estimations by means of dependency-based or of thematic roles-based DSMs. However, these semantic spaces are built similarly to traditional DSMs as they split verb arguments into separate vector dimensions. By using syntactic-semantic links, they encode the relation between an event and each of its participants, but they do not encode directly the relation between participants co-occurring in the same event. 
% QUI AGGIUNGI UN ESEMPIO? O FORSE MEGLIO DOPO?

Another trend of studies in the NLP community aimed at the introduction of richer contextual features in DSMs, mostly based on word windows. The first example was the composite-feature model by Ruiz-Casado et al. \shortcite{ruiz2005using}, who extracted word windows through a Web Search engine. A composite feature for the target word \textit{watches} is \textit{Alicia always \_\_\_\_ romantic movies}, extracted from the sentence \textit{I heard that Alicia always watches romantic movies with Antony} (the placeholder represents the target position). Thanks to this approach, Ruiz-Casado and colleagues achieved 82.50 in the TOEFL synonym detection test, outperforming Latent Semantic Analysis (LSA; see Landauer et al. \shortcite{landauer1998introduction}) and several other methods.

Agirre et al. \shortcite{agirre2009study} adopted an analogous approach, relying on a huge learning corpus (1.6 Teraword) to build composite-feature vectors. Their model outperformed a traditional DSM on the similarity subset of the WordSim-353 test set \cite{finkelstein2001placing}.

Melamud et al. \shortcite{melamud2014probabilistic} introduced a probabilistic similarity scheme for modeling the so-called joint context. By making use of the Kneser-Ney language model \cite{kneser1995improved} and of a probabilistic distributional measure, they were able to overcome data sparsity, outperforming a wide variety of DSMs on two similarity tasks, evaluated on VerbSim \cite{yang2006verb} and on a set of 1,000 verbs extracted from WordNet \cite{fellbaum1998wordnet}. On the basis of their results, the authors claimed that composite-feature models are particularly advantageous for measuring verb similarity.

\section{Syntactic joint contexts}

A joint context, as defined in Melamud et al. \shortcite{melamud2014probabilistic}, is a word window of order \textit{n} around a target word. The target is replaced by a placeholder, and the value of the feature for a word \textit{w} is the probability of \textit{w} to fill the placeholder position. Assuming \textit{n}=3, a word like \textit{love} would be represented by a collection of contexts such as \textit{the new students  \_\_\_\_ the school campus}. Such representation introduces data sparseness, which has been addressed by previous studies either by adopting huge corpora or by relying on n-gram language models to approximate the probabilities of long sequences of words.

However, features based on word windows do not guarantee to include all the most salient event participants. Moreover, they could include unrelated words, also differentiating contexts describing the same event (e.g. consider \textit{Luis \_\_\_\_ the red ball} and \textit{Luis \_\_\_\_ the blue ball}).

For these reasons, we introduce the notion of \textit{syntactic joint contexts}, further abstracting from linear word windows by using dependencies. %Our assumption is that knowledge about typical event participants can be inferred by observing the most frequent argument combinations, so this is the type of information that we want to include in our features. 
Each feature of the word vector, in our view, should correspond to a typical verb-argument combination, as an approximation to our knowledge about typical event participants. In the present study, we are focusing on verbs because verb meaning is highly context sensitive and include information about complex argument configurations. Therefore, verb representation should benefit more from the introduction of joint features \cite{melamud2014probabilistic}.

The procedure for defining of our representations is the following:

\begin{itemize}
\item we extract a list of verb-argument dependencies from a parsed corpus, and for each target verb we extract all the direct dependencies from the sentence of occurrence. For instance, in \textit{Finally, the dictator acknowledged his failure}, we will have: target = 'acknowledge-v'; subject = 'dictator-n'; and object = 'failure-n'.

\item for each sentence, we generate a joint context feature by joining all the dependencies for the grammatical relations of interest. From the example above, we would generate the feature \textit{dictator-n.subj+\_\_\_\_+failure-n.obj}.
\end{itemize}

For our experiments, the grammatical relations that we used are \textit{subject}, \textit{object} and \textit{complement}, where \textit{complement} is a generic relation grouping together all dependencies introduced by a preposition.
Our distributional representation for a target word is a vector of syntatic joint contexts. For instance, the word vector for the verb \textit{to begin} would include features like \{\textit{jury-n.subj+\_\_\_\_+deliberation-n.obj}, \textit{operation-n.subj+\_\_\_\_+on-i\_thursday-n.comp}, \textit{recruit-n.subj+\_\_\_\_+training-n.obj+on-i\_street-n.comp} ...\}. The value of each joint feature will be the frequency of occurrence of the target verb with the corresponding argument combination, possibly weighted by some statistical association measure.
% inserisci un esempio con la relazione complemento

\section{Evaluation}

\subsection{Corpus and DSMs}

We trained our DSMs on the RCV1 corpus, which contains approximately 150 million words \cite{lewis2004rcv1}. The corpus was tagged with the tagger described in Dell'Orletta \shortcite{dell2009ensemble} and dependency-parsed with DeSR \cite{attardi2009accurate}. RCV1 was chosen for two reasons: i) to show that our joint context-based representation can deal with data sparseness even with a training corpus of limited size; ii) to allow a comparison with the results reported by Melamud et al. \shortcite{melamud2014probabilistic}. 
% inverti e approfondisci

All DSMs adopt Positive Pointwise Mutual Information (PPMI; Church and Hanks \shortcite{church1990word}) as a context weighting scheme and vary according to three main parameters: i) type of contexts; ii) number of dimensions; iii) application of Singular Value Decomposition (SVD; see Landauer et al. \shortcite{landauer1998introduction}).

For what concerns the first parameter, we developed three types of DSMs: a) traditional bag-of-words DSMs, where the features are content words co-occurring with the target in a window of width 2; b) dependency-based DSMs, where the features are words in a direct dependency relation with the target; c) joint context-based DSMs, using the joint features described in the previous section. 
The second parameter refers instead to the number of contexts that have been used as vector dimensions. Several values were explored (i.e. 10K, 50K and 100K), selecting the contexts according to their frequency. Finally, the third parameter concerns the application of SVD to reduce the matrix. We report only the results for a number \textit{k} of latent dimensions ranging from 200 to 400, since the performance drops significantly out of this interval.

\subsection{Similarity Measures}

As a similarity measure, we used vector cosine, which is by far the most popular in the existing literature \cite{turney2010frequency}.
Melamud et al. \shortcite{melamud2014probabilistic} have proposed the Probabilistic Distributional Similarity (PDS), based on the intuition that two words, $w_1$ and $w_2$, are similar if they are likely to occur in each other's contexts. PDS assigns a high similarity score when both $p(w_1 |$ contexts of $w_2)$ and $p(w_2 |$ contexts of $w_1)$ are high. We tried to test variations of this measure with our representation, but we were not able to achieve satisfying results. Therefore, we report here only the scores with the cosine.

\subsection{Datasets}
\label{ssec:datasets}

The DSMs are evaluated on two test sets: VerbSim \cite{yang2006verb} and the verb subset of SimLex-999 \cite{hill2015simlex}. The former includes 130 verb pairs, while the latter includes 222 verb pairs. %We focus on verb similarity because verb meaning is highly context sensitive and, therefore, verb representation should benefit more from the introduction of joint features \cite{melamud2014probabilistic}.

Both datasets are annotated with similarity judgements, so we measured the Spearman correlation between them and the scores assigned by the model.
The VerbSim dataset allows for comparison with Melamud et al. \shortcite{melamud2014probabilistic}, since they also evaluated their model on this test set, achieving a Spearman correlation score of 0.616 and outperforming all the baseline methods.

The verb subset of SimLex-999, at the best of our knowledge, has never been used as a benchmark dataset for verb similarity.
The SimLex dataset is known for being quite challenging: as reported by Hill et al. \shortcite{hill2015simlex}, the average performances of similarity models on this dataset are much lower than on alternative benchmarks like WordSim \cite{finkelstein2001placing} and MEN \cite{bruni2014multimodal}.

We exclude from the evaluation datasets all the target words occurring less than 100 times in our corpus. Consequently, we cover 107 pairs in the VerbSim dataset (82.3, the same of Melamud et al. \shortcite{melamud2014probabilistic}) and 214 pairs in the SimLex verbs dataset (96.3).

\subsection{Results}

Table 1 reports the Spearman correlation scores for the vector cosine on our DSMs. At a glance, we can notice the discrepancy between the results obtained in the two datasets, as SimLex verbs confirms to be very difficult to model. We can also recognize a trend related to the number of contexts, as the performance tends to improve when more contexts are taken into account (with some exceptions). Single dependencies and joint contexts perform very similarly, and no one has a clear edge on the other. Both of them outperform the bag-of-words model on the VerbSim dataset by a nice margin, whereas the scores of all the model types are pretty much the same on SimLex verbs.
Finally, it is noteworthy that the score obtained on VerbSim by the joint context model with 100K dimensions goes very close to the result reported by Melamud et al. \shortcite{melamud2014probabilistic} (0.616).

Table 2 and Table 3 report the results of the models with SVD reduction. Independently of the number of dimensions \textit{k}, the joint contexts almost always outperform the other model types. Overall, the performance of the joint contexts seems to be more stable across several parameter configurations, whereas bag-of-words and single dependencies are subject to bigger drops. Exceptions can be noticed only for the VerbSim dataset, and only with a low number of dimensions. Finally, the correlation coefficients for the two datasets seem to follow different trends, as the models with a higher number of contexts perform better on SimLex verbs, while the opposite is true for the VerbSim dataset.

On the VerbSim dataset, both single dependencies and joint contexts have again a clear advantage over bag-of-words representations Although they achieve a similar performance with 10K contexts, the correlation scores of the former decrease more quickly as the number of contexts increases, while the latter are more stable. Moreover, joint contexts are able to outperform single dependencies.\\
On SimLex verbs, all the models are very close and -- differently from the previous dataset -- the higher-dimensional DSMs are the better performing ones. Though differences are not statistically significant, joint context are able to achieve top scores over the other models.\footnote{p-values computed with Fisher's r-to-z transformation comparing correlation coefficients between the joint context-DSMs and the other models on the same parameter settings.}

More in general, the best results are obtained with SVD reduction and \textit{k}=200. The joint context-based DSM with 10K dimensions and \textit{k} = 200 achieves 0.65, which is above the result of Melamud et al. \shortcite{melamud2014probabilistic}, although the difference between the two correlation scores is not significant. As for SimLex verbs, the best result (0.283) is obtained by the joint context DSM with 100K dimensions and \textit{k} = 200.

\begin{table}[ht]
\small
\centering
\label{my-label}
\begin{tabular}{|c|c|c|}
\hline
\textbf{Model} & \textbf{VerbSim} & \textbf{SimLex verbs} \\ \hline
\textbf{Bag-of-Words-10K} & 0.385						 & 0.085										\\ \hline
\textbf{Single - 10k}     & 0.561            & 0.090                    \\ \hline
\textbf{Joint - 10k}      & 0.568            & 0.105                    \\ \hline
\textbf{Bag-of-Words-50K} & 0.478						 & 0.095										\\ \hline
\textbf{Single - 50k}     & 0.592            & 0.115                    \\ \hline
\textbf{Joint - 50k}      & 0.592            & 0.105                    \\ \hline
\textbf{Bag-of-Words-100K} & 0.488						 & 0.114										\\ \hline
\textbf{Single - 100k}    & 0.587            & \textbf{0.132}           \\ \hline
\textbf{Joint - 100k}     & \textbf{0.607}   & 0.114                    \\ \hline
\end{tabular}
\caption{Spearman correlation scores for VerbSim and for the verb subset of SimLex-999. Each model is identified by the type and by the number of features of the semantic space.}
\end{table}

\begin{table}[ht]
\small
\centering
\label{my-labelsecond}
\begin{tabular}{|c|c|c|c|}
\hline
	\textbf{Model} & \textbf{k = 200} & \textbf{k = 300} & \textbf{k = 400} \\ \hline
	\textbf{Bag-of-Words-10K} & 0.457 & 0.445	& 0.483 \\ \hline
	\textbf{Single - 10k} & 0.623 & 0.647 & 0.641  \\ \hline
	\textbf{Joint - 10k} & \textbf{0.650} & 0.636 & 0.635  \\ \hline
	\textbf{Bag-of-Words-50K} & 0.44 & 0.453 & 0.407	 \\ \hline
	\textbf{Single - 50k} & 0.492 & 0.486 & 0.534  \\ \hline
	\textbf{Joint - 50k} & 0.571 & 0.591 & 0.613  \\ \hline
	\textbf{Bag-of-Words-100K} & 0.335 & 0.324 & 0.322 \\ \hline
	\textbf{Single - 100k} & 0.431 & 0.413 & 0.456  \\ \hline
	\textbf{Joint - 100k} & 0.495 & 0.518 & 0.507  \\ \hline
\end{tabular}
\caption{Spearman correlation scores for VerbSim, after the application of SVD with different values of \textit{k}.} %The first column states whether the DSM is implementing single dependencies or joint contexts and the number of contexts taken into account.}
\end{table}

\begin{table}[ht]
\small
\centering
\label{my-labelthird}
\begin{tabular}{|c|c|c|c|}
\hline
	\textbf{Model} & \textbf{k = 200} & \textbf{k = 300} & \textbf{k = 400} \\ \hline
	\textbf{Bag-of-Words-10K} & 0.127 & 0.113 &	0.111	\\ \hline
	\textbf{Single - 10k} & 0.168 & 0.172 & 0.165  \\ \hline
	\textbf{Joint - 10k} & 0.190 & 0.177 & 0.181  \\ \hline
	\textbf{Bag-of-Words-50K} & 0.196 & 0.191 & 0.21  \\ \hline
	\textbf{Single - 50k} & 0.218 & 0.228 & 0.222  \\ \hline
	\textbf{Joint - 50k} & 0.256 & 0.250 & 0.227  \\ \hline
	\textbf{Bag-of-Words-100K} & 0.222 & 0.18 & 0.16 \\ \hline
	\textbf{Single - 100k} & 0.225 & 0.218 & 0.199  \\ \hline
	\textbf{Joint - 100k} & \textbf{0.283} & 0.256 & 0.222  \\ \hline
\end{tabular}
\caption{Spearman correlation scores for the verb subset of SimLex-999, after the application of SVD with different values of \textit{k}.} 
\end{table}

\subsection{Conclusions}

In this paper, we have presented our proposal for a new type of vector representation based on joint features, which should emulate more closely the general knowledge about event participants that seems to be the organizing principle of our mental lexicon. A core issue of previous studies was the data sparseness challenge, and we coped with it by means of a more abstract, syntactic notion of joint context.

The models using joint dependencies were able at least to perform comparably to traditional, dependency-based DSMs. In our experiments, they even achieved the best correlation scores across several parameter settings, especially after the application of SVD.
We want to emphasize that previous works such as Agirre et al. \shortcite{agirre2009study} already showed that large word windows can have a higher discriminative power than indipendent features, but they did it by using a huge training corpus. In our study, joint context-based representations derived from a small corpus such as RCV1 are already showing competitive performances. This result strengthens our belief that dependencies are a possible solution for the data sparsity problem of joint feature spaces. 

We also believe that verb similarity might not be the best task to show the usefulness of joint contexts for semantic representation. The main goal of the present paper was to show that joint contexts are a viable option to exploit the full potential of distributional information. Our successful tests on verb similarity prove that syntactic joint contexts do not suffer of data sparsity and are also able to beat other types of representations based on independent word features. Moreover, syntactic joint contexts are much simpler and more competitive with respect to window-based ones. \\
The good performance in the verb similarity task motivates us to further test syntactic joint contexts on a larger range of tasks, such as word sense disambiguation, textual entailment and classification of semantic relations, so that they can unleash their full potential. Moreover, our proposal opens interesting perspectives for computational psycholinguistics, especially for modeling those semantic phenomena that are inherently related to the activation of event knowledge (e.g. thematic fit).

\section*{Acknowledgments}

This paper is partially supported by HK PhD Fellowship Scheme, under PF12-13656. Emmanuele Chersoni's research is funded by a grant of the University Foundation A*MIDEX.

%\bibliographystyle{emnlp2016}
%\bibliography{emnlp2016}

\begin{thebibliography}{}
\bibitem[\protect\citename{Agirre \bgroup et al.\egroup }2009]{agirre2009study}
Eneko Agirre, Enrique Alfonseca, Keith Hall, Jana Kravalova, Marius
  Pa{\c{s}}ca, and Aitor Soroa.
\newblock 2009.
\newblock A study on similarity and relatedness using distributional and
  wordnet-based approaches.
\newblock In {\em Proceedings of the 2009 conference of the NAACL-HLT}, pages 19--27. Association for Computational Linguistics.

\bibitem[\protect\citename{Attardi \bgroup et al.\egroup
  }2009]{attardi2009accurate}
Giuseppe Attardi, Felice Dell'Orletta, Maria Simi, and Joseph Turian.
\newblock 2009.
\newblock Accurate dependency parsing with a stacked multilayer perceptron.
\newblock In {\em Proceedings of EVALITA}, 9.

\bibitem[\protect\citename{Baroni and Lenci}2010]{baroni2010distributional}
Marco Baroni and Alessandro Lenci.
\newblock 2010.
\newblock Distributional memory: A general framework for corpus-based
  semantics.
\newblock {\em Computational Linguistics}, 36(4):673--721.

\bibitem[\protect\citename{Bicknell \bgroup et al.\egroup
  }2010]{bicknell2010effects}
Klinton Bicknell, Jeffrey~L Elman, Mary Hare, Ken McRae, and Marta Kutas.
\newblock 2010.
\newblock Effects of event knowledge in processing verbal arguments.
\newblock {\em Journal of Memory and Language}, 63(4):489--505.

\bibitem[\protect\citename{Bruni \bgroup et al.\egroup
  }2014]{bruni2014multimodal}
Elia Bruni, Nam-Khanh Tran, and Marco Baroni.
\newblock 2014.
\newblock Multimodal distributional semantics.
\newblock {\em J. Artif. Intell. Res.(JAIR)}, 49(1-47).

\bibitem[\protect\citename{Church and Hanks}1990]{church1990word}
Kenneth~Ward Church and Patrick Hanks.
\newblock 1990.
\newblock Word association norms, mutual information, and lexicography.
\newblock {\em Computational linguistics}, 16(1):22--29.

\bibitem[\protect\citename{Dell'Orletta}2009]{dell2009ensemble}
Felice Dell'Orletta.
\newblock 2009.
\newblock Ensemble system for part-of-speech tagging.
\newblock In {\em Proceedings of EVALITA}, 9.

\bibitem[\protect\citename{Fellbaum}1998]{fellbaum1998wordnet}
Christiane Fellbaum.
\newblock 1998.
\newblock {\em WordNet}.
\newblock Wiley Online Library.

\bibitem[\protect\citename{Finkelstein \bgroup et al.\egroup
  }2001]{finkelstein2001placing}
Lev Finkelstein, Evgeniy Gabrilovich, Yossi Matias, Ehud Rivlin, Zach Solan,
  Gadi Wolfman, and Eytan Ruppin.
\newblock 2001.
\newblock Placing search in context: The concept revisited.
\newblock In {\em Proceedings of the 10th international conference on World
  Wide Web}, pages 406--414. ACM.

\bibitem[\protect\citename{Greenberg \bgroup et al.\egroup
  }2015]{greenberg2015improving}
Clayton Greenberg, Asad Sayeed, and Vera Demberg.
\newblock 2015.
\newblock Improving unsupervised vector-space thematic fit evaluation via
  role-filler prototype clustering.
\newblock In {\em Proceedings of the 2015 conference of the NAACL-HLT, Denver, USA}.

\bibitem[\protect\citename{Hare \bgroup et al.\egroup
  }2009]{hare2009activating}
Mary Hare, Michael Jones, Caroline Thomson, Sarah Kelly, and Ken McRae.
\newblock 2009.
\newblock Activating event knowledge.
\newblock {\em Cognition}, 111(2):151--167.

\bibitem[\protect\citename{Harris}1954]{harris1954distributional}
Zellig~S Harris.
\newblock 1954.
\newblock Distributional structure.
\newblock {\em Word}, 10(2-3):146--162.

\bibitem[\protect\citename{Hill \bgroup et al.\egroup }2015]{hill2015simlex}
Felix Hill, Roi Reichart, and Anna Korhonen.
\newblock 2015.
\newblock Simlex-999: Evaluating semantic models with (genuine) similarity
  estimation.
\newblock {\em Computational Linguistics}.

\bibitem[\protect\citename{Kneser and Ney}1995]{kneser1995improved}
Reinhard Kneser and Hermann Ney.
\newblock 1995.
\newblock Improved backing-off for m-gram language modeling.
\newblock In {\em Acoustics, Speech, and Signal Processing, 1995. ICASSP-95.,
  1995 International Conference on}, volume~1, pages 181--184. IEEE.

\bibitem[\protect\citename{Landauer \bgroup et al.\egroup
  }1998]{landauer1998introduction}
Thomas~K Landauer, Peter~W Foltz, and Darrell Laham.
\newblock 1998.
\newblock An introduction to latent semantic analysis.
\newblock {\em Discourse processes}, 25(2-3):259--284.

\bibitem[\protect\citename{Lenci}2011]{lenci2011composing}
Alessandro Lenci.
\newblock 2011.
\newblock Composing and updating verb argument expectations: A distributional
  semantic model.
\newblock In {\em Proceedings of the 2nd Workshop on Cognitive Modeling and
  Computational Linguistics}, pages 58--66. Association for Computational
  Linguistics.

\bibitem[\protect\citename{Lewis \bgroup et al.\egroup }2004]{lewis2004rcv1}
David~D Lewis, Yiming Yang, Tony~G Rose, and Fan Li.
\newblock 2004.
\newblock Rcv1: A new benchmark collection for text categorization research.
\newblock {\em The Journal of Machine Learning Research}, 5:361--397.

\bibitem[\protect\citename{McRae \bgroup et al.\egroup
  }1998]{mcrae1998modeling}
Ken McRae, Michael~J Spivey-Knowlton, and Michael~K Tanenhaus.
\newblock 1998.
\newblock Modeling the influence of thematic fit (and other constraints) in
  on-line sentence comprehension.
\newblock {\em Journal of Memory and Language}, 38(3):283--312.

\bibitem[\protect\citename{McRae \bgroup et al.\egroup }2005]{mcrae2005basis}
Ken McRae, Mary Hare, Jeffrey~L Elman, and Todd Ferretti.
\newblock 2005.
\newblock A basis for generating expectancies for verbs from nouns.
\newblock {\em Memory \& Cognition}, 33(7):1174--1184.

\bibitem[\protect\citename{Melamud \bgroup et al.\egroup
  }2014]{melamud2014probabilistic}
Oren Melamud, Ido Dagan, Jacob Goldberger, Idan Szpektor, and Deniz Yuret.
\newblock 2014.
\newblock Probabilistic modeling of joint-context in distributional similarity.
\newblock In {\em CoNLL}, pages 181--190.

\bibitem[\protect\citename{Mikolov \bgroup et al.\egroup
  }2013]{mikolov2013efficient}
Tomas Mikolov, Kai Chen, Greg Corrado, and Jeffrey Dean.
\newblock 2013.
\newblock Efficient estimation of word representations in vector space.
\newblock {\em arXiv preprint arXiv:1301.3781}.

\bibitem[\protect\citename{Ruiz-Casado \bgroup et al.\egroup
  }2005]{ruiz2005using}
Maria Ruiz-Casado, Enrique Alfonseca, and Pablo Castells.
\newblock 2005.
\newblock Using context-window overlapping in synonym discovery and ontology
  extension.
\newblock In {\em Proceedings of RANLP}, pages 1--7.

\bibitem[\protect\citename{Sahlgren}2008]{sahlgren2008distributional}
Magnus Sahlgren.
\newblock 2008.
\newblock The distributional hypothesis.
\newblock {\em Italian Journal of Linguistics}, 20(1):33--54.

\bibitem[\protect\citename{Santus et al.}2016]{santus2016testing}
Enrico Santus, Emmanuele Chersoni, Alessandro Lenci, Chu-Ren Huang, and Philippe Blache.
\newblock 2016.
\newblock Testing APSyn against Vector Cosine on Similarity Estimation.
\newblock In {\em Proceedings of the Pacific Asia Conference on Language, Information and Computing (PACLIC 30)}.

\bibitem[\protect\citename{Sayeed and Demberg}2014]{sayeed2014combining}
Asad Sayeed and Vera Demberg.
\newblock 2014.
\newblock Combining unsupervised syntactic and semantic models of thematic fit.
\newblock In {\em Proceedings of the first Italian Conference on Computational
  Linguistics (CLiC-it 2014)}.

\bibitem[\protect\citename{Turney \bgroup et al.\egroup
  }2010]{turney2010frequency}
Peter~D Turney, Patrick Pantel, et~al.
\newblock 2010.
\newblock From frequency to meaning: Vector space models of semantics.
\newblock {\em Journal of artificial intelligence research}, 37(1):141--188.

\bibitem[\protect\citename{Weeds \bgroup et al.\egroup
  }2004]{weeds2004characterising}
Julie Weeds, David Weir, and Diana McCarthy.
\newblock 2004.
\newblock Characterising measures of lexical distributional similarity.
\newblock In {\em Proceedings of the 20th international conference on
  Computational Linguistics}, page 1015. Association for Computational
  Linguistics.

\bibitem[\protect\citename{Yang and Powers}2006]{yang2006verb}
Dongqiang Yang and David~MW Powers.
\newblock 2006.
\newblock {\em Verb similarity on the taxonomy of WordNet}.
\newblock Masaryk University.

\end{thebibliography}

\end{document}